%% file: paper.tex
\documentclass{article}

\usepackage{microtype}
\usepackage{graphicx}
\usepackage{subfigure}
\usepackage{booktabs} 

\usepackage{breakurl}
\usepackage[breaklinks]{hyperref}

\usepackage[accepted]{mlsys2024}

\mlsystitlerunning{\Sys: Optimizing Auto-batching of Dynamic Deep
  Learning at Compile Time}

\usepackage{amsmath}
\usepackage{amssymb}
\usepackage{mathtools}
\usepackage{amsthm}

\usepackage[capitalize,noabbrev]{cleveref}

\theoremstyle{plain}

\theoremstyle{definition}

\theoremstyle{remark}

\usepackage[textsize=tiny]{todonotes}

\usepackage{hyperref}

\usepackage[flushleft]{threeparttable}
\usepackage{enumitem}
\usepackage[frozencache,newfloat]{minted}
\usepackage{listings}
\usepackage{multirow}
\usepackage{wrapfig}
\usepackage{amsmath,graphicx}
\usepackage{tikz}
\usepackage{array}

\include{defs}

\begin{document}

\twocolumn[ \mlsystitle{\Sys: Optimizing Auto-batching of\\Dynamic
    Deep Learning at Compile Time}



\mlsyssetsymbol{equal}{*}

\begin{mlsysauthorlist}
\mlsysauthor{Pratik Fegade}{cmu}
\mlsysauthor{Tianqi Chen}{cmu,octai}
\mlsysauthor{Phillip B. Gibbons}{cmu}
\mlsysauthor{Todd C. Mowry}{cmu}
\end{mlsysauthorlist}

\mlsysaffiliation{cmu}{Carnegie Mellon University, Pittsburgh, USA}
\mlsysaffiliation{octai}{OctoAI}

\mlsyscorrespondingauthor{Pratik Fegade}{pratikfegade@gmail.com}

\mlsyskeywords{control flow, dynamic models, compilers, batching,
  auto-batching, dynamic batching}

\vskip 0.3in

\begin{abstract}
  \input{src/abstract.tex}
\end{abstract}
]



\printAffiliationsAndNotice{}  
.

\input{src/introduction.tex}
\input{src/background.tex}
\input{src/overview.tex}
\input{src/hybrid_analysis.tex}
\input{src/kernels.tex}
\input{src/implementation.tex}
\input{src/evaluation.tex}
\input{src/related_work.tex}
\input{src/conclusion.tex}
\input{src/acks.tex}

\bibliography{paper}
\bibliographystyle{mlsys2024}

\newpage
\begin{appendix}
\input{src/appendix.tex}
\end{appendix}
\end{document}

%% file: defs.tex
\newcommand{\Sys}{\textsc{ACRoBat}}

\newcommand*\circled[1]{\tikz[baseline=(char.base)]{
            \node[shape=circle,draw,thick,inner sep=1pt, scale = 0.8] (char) {#1};}}

\newcolumntype{L}[1]{>{\raggedright\let\newline\\\arraybackslash\hspace{0pt}}m{#1}}
\newcolumntype{C}[1]{>{\centering\let\newline\\\arraybackslash\hspace{0pt}}m{#1}}

\newif\ifminted
\mintedtrue

\makeatletter\let\expandableinput\@@input\makeatother

%% file: src/abstract.tex

Dynamic control flow is an important technique often used to design
expressive and efficient deep learning computations for applications
such as text parsing, machine translation, exiting early out of deep
models and so on. The control flow divergence resulting from dynamic
control flow makes batching, an important optimization enabling high
throughput and hardware utilization, difficult to perform manually. In
this paper, we present \Sys, a framework that enables efficient
automatic batching for dynamic deep learning computations by
performing hybrid static+dynamic compiler optimizations and end-to-end
tensor code generation. \Sys~performs up to 8.5$\times$~better than
DyNet, a state-of-the-art framework for automatic batching, on an
Nvidia GeForce GPU.

%% file: src/introduction.tex
\section{Introduction}\label{sec:intro}
Deep Learning (DL) has come to play an increasing role in a wide range
of applications in the recent years. As their applications have become
more and more complex, DL models themselves have increased in size and
complexity. For inference serving as well as for training, these
models place extreme demands on DL systems and hardware today.

An important source of complexity in DL computations is the use of
dynamic control flow as part of execution. Unlike a static
feed-forward model computation, the execution of a computation with
dynamic control flow, or a \emph{dynamic computation} can differ
across different inputs to the model. Among other applications, this
property has been used effectively to (1) model structured data such
as parse trees~\cite{treelstm, mvrnn} and images~\cite{dagrnn}, (2)
perform better quality machine translations and text parsing by
employing beam search~\cite{beam_search, pharoah,
  beam_search_parsing}, and (3) exit early out of
convolutional~\cite{early_exit_cnn1, early_exit_cnn2} and
transformer~\cite{deebert, depth_adaptive_transformer} models for
reduced inference latency. The adaptability afforded by dynamic
control flow is thus useful in a variety of situations.

Batching is an important optimization that improves the throughput and
hardware utilization during training and inference of a DL
model. While straightforward for static DL computations, the presence
of control flow divergence in dynamic computations makes manual
batching difficult and error-prone. Thus, there has been significant
past effort on performing automatic batching, or \emph{auto-batching},
for dynamic DL computations. In order to handle the lack of execution
knowledge of a dynamic computation during compilation, past works
usually either (1) heavily rely on dynamic analyses, enabling them to
handle general dynamic control flow~\cite{dynet2, tffold}, or (2) are
specialized for specific control flow patterns or models, thus relying
more on static analyses~\cite{cavs, cortex}. The former frameworks
often \emph{incur high execution overheads} caused by dynamic
analysis, while the latter ones \emph{lack the generality} to support
the wide range of existing and future control flow patterns in DL
computations.


Further, past work often \emph{heavily relies on vendor libraries}
such as cuDNN~\cite{cudnn} and oneAPI~\cite{onednn}. However, as
implementing vendor libraries is an intensive process, they usually
only implement commonly used, standard tensor operators. Further, as
these kernels are optimized in isolation, without any contextual
knowledge about the larger application they are used in, important
optimizations such as kernel fusion can no longer be performed.

In order to overcome these limitations of past work, we propose
\Sys\footnote{\textbf{A}utomated \textbf{C}ompiler and
\textbf{R}untime-\textbf{o}ptimized \textbf{Bat}ching}, an
auto-batching framework for dynamic DL computations which relies on
novel \emph{hybrid static+dynamic optimizations} and \emph{end-to-end
tensor kernel compilation}. Our main insight in designing \Sys~is that
despite the lack of perfect execution knowledge during compilation for
dynamic models, the compiler can often perform static analysis and
optimizations to aid the dynamic analysis. This reduces execution
overheads while effectively exploiting parallelism in the input
computation. \Sys~relies on traditional compiler techniques such as
context-sensitivity~\cite{dragon_book} and taint analysis as well as
on minimal user annotations to enable such static analysis. Further,
\Sys's end-to-end tensor kernel generation enables it to automatically
generate kernels optimized and specialized to the larger computation
again using static analysis to identify and exploit data reuse
opportunities (as we see in~\S\ref{sec:kernels}). \Sys's generality
allows one to express a wide variety of control flow patterns, ranging
from simple conditional statements to complex recursive computations
using a simple high-level language. Table~\ref{table:comp} provides a
qualitative comparison of \Sys~with related work.

\newcommand{\gyes}[0]{\textcolor{green}{Yes}}
\newcommand{\rno}[0]{\textcolor{red}{No}}
\newcommand{\ghigh}[0]{\textcolor{green}{High}}
\newcommand{\gnone}[0]{\textcolor{green}{None}}
\newcommand{\ymid}[0]{\textcolor{orange}{Mid}}
\newcommand{\rlow}[0]{\textcolor{red}{Low}}
\newcommand{\rhigh}[0]{\textcolor{red}{High}}
\newcommand{\glow}[0]{\textcolor{green}{Low}}
\newcommand{\rstatic}[0]{\textcolor{red}{Static only}}
\newcommand{\rdynamic}[0]{\textcolor{red}{Dyn. only}}
\newcommand{\ghybrid}[0]{\textcolor{green}{Hybrid}}

\begin{table}[t]
  \vspace{-4mm}
\caption{Comparison between \Sys~and other solutions for auto-batching
  dynamic DL computations. Purely static or dynamic approaches can be
  overly conservative, or have high overheads respectively, unlike
  \Sys's hybrid analysis.}
  \vspace{-1mm}
\label{table:comp}
\begin{center}
\begin{scriptsize}
  \addtolength{\tabcolsep}{-5pt}
  \addtolength\extrarowheight{-0.5pt}
\resizebox{0.99\columnwidth}{!}{%
\begin{tabular}{C{2.25cm}|C{1.15cm}C{1.3cm}C{1.15cm}C{1.3cm}|C{1.15cm}}
\toprule
Framework              & PyTorch   & DyNet     & Cortex   & TFFold    & \Sys           \\
\midrule
Auto-batch support     & \rno      & \gyes     & \gyes    & \gyes     & \gyes          \\
Auto-batch analysis    & -         & \rdynamic & \rstatic & \rdynamic & \ghybrid       \\

Vendor library use     & \rhigh    & \rhigh    & \gnone   & \rhigh    & \gnone         \\

Generality             & \ghigh    & \ghigh    & \rlow    & \ymid     & \ghigh         \\
User impl. effort      & \glow     & \glow     & \rhigh   & \glow     & \glow          \\
Performance            & \rlow     & \rlow     & \ghigh   & \rlow     & \ghigh         \\
\bottomrule
\end{tabular}%
}
\addtolength\extrarowheight{0.5pt}
\addtolength{\tabcolsep}{4pt}
\end{scriptsize}
\end{center}
\vspace{-3mm}
\end{table}

In short, this paper makes the following contributions:
\begin{enumerate}[topsep=-0.7ex, leftmargin=1.2em, itemsep=-0.5ex]
  \item We survey and characterize the dynamic control flow structures
    found in different DL computations.
  \item Employing novel hybrid static+dynamic optimizations and
    automated end-to-end kernel code generation, we design \Sys, an
    auto-batching framework for dynamic computations. This design
    allows us to reduce execution overheads and to generate efficient
    tensor kernels that effectively exploit data reuse
    opportunities. In developing these optimizations, we heavily rely
    on traditional compilation techniques.
  \item We prototype \Sys, evaluate it against state-of-the-art deep
    learning frameworks~\cite{cavs, dynet1,pytorch} and report
    significant performance gains on Nvidia GPUs.
\end{enumerate}

%% file: src/background.tex
\definecolor{darkgreen}{RGB}{0,160,0}
\def\checkmark{\tikz\fill[scale=0.3,darkgreen](0,.35) -- (.25,0) -- (1,.7) -- (.25,.15) -- cycle;}

\begin{table}[t]
\vspace{-3mm}
  \caption{Control flow properties found in DL computations. Legend:
    ITE: iterative control flow, REC: recursive control flow, TDC:
    model exhibits tensor-dependent control flow (where control flow
    decisions are predicated on values on intermediate tensors), IP:
    computation exhibits high instance parallelism, ICF: model inference
    exhibits control flow, TCF: model training exhibits control flow.}
\label{table:dynamism}
\begin{center}
\begin{scriptsize}
\addtolength{\tabcolsep}{-4pt}
\addtolength\extrarowheight{-0.5pt}
\resizebox{0.99\columnwidth}{!}{%
\begin{tabular}{p{5cm}|cccccc}
\toprule
    Deep Learning Computations                                  & ITE & REC & TDC & IP & ICF & TCF  \\
    \midrule

    RNN~\cite{rnn}, LSTM~\cite{lstm}, GRU~\cite{gru}, GraphRNN~\cite{graph_rnn} & \checkmark &  &  &  & \checkmark & \checkmark  \\ \hline
    Speculative decoding for transformers~\cite{spec_decode} & \checkmark &      & \checkmark &    \checkmark  & \checkmark &  \\ \hline
    DIORA~\cite{diora}, Chinese Segmentation~\cite{ch_seg} & \checkmark &  &  & \checkmark & \checkmark & \checkmark  \\ \hline
    DAG-RNN~\cite{dagrnn}, TreeLSTM~\cite{treelstm}, MV-RNN~\cite{mvrnn} &  & \checkmark &  & \checkmark & \checkmark & \checkmark  \\ \hline
    StackLSTM~\cite{stacklstm} & \checkmark &  & \checkmark &  & \checkmark & \checkmark  \\ \hline
    Beam search~\cite{beam_search} with LSTM & \checkmark &  & \checkmark & \checkmark & \checkmark &   \\ \hline
    Mixture-of-experts~\cite{moe1, moe2, switch_trans} &  &  & \checkmark &  & \checkmark & \checkmark  \\ \hline
    Early exit models~\cite{early_exit_cnn1, early_exit_cnn2, depth_adaptive_transformer} &  &  & \checkmark &  & \checkmark &   \\ \hline
    Tree-to-tree NN~\cite{tree2tree}, Doubly Recurrent NN~\cite{drnn} &  & \checkmark & \checkmark & \checkmark & \checkmark & \checkmark  \\ \hline
    R-CNN~\cite{rcnn}, Fast R-CNN~\cite{fast_rcnn} & \checkmark &  & \checkmark & \checkmark & \checkmark & \checkmark  \\
\bottomrule
\end{tabular}%
}
\addtolength\extrarowheight{0.5pt}
\addtolength{\tabcolsep}{4pt}
\end{scriptsize}
\end{center}
\vspace{-5mm}
\end{table}

\section{Background}\label{sec:background}

\subsection{Dynamic Control Flow in DL computations}\label{sec:survey}
In this section, we take a look at the different kinds of control flow
dynamism present in various DL computations in the context of the
auto-batching problem. This will inform how we design a system to
exploit parallelism across tensor operators in the batched execution
of dynamic DL computation.

Note that given a computation involving control flow, there are often
multiple ways to implement it. We consider the most natural way to
implement a given computation. For example, a top-down tree traversal
can be implemented as a breadth-first traversal (BFS) or a depth-first
traversal (DFS). While a BFS traversal can be more efficient, the
DFS-based traversal is more natural to implement. The discussion below
is also summarized in Table~\ref{table:dynamism}.

\noindent\textbf{Control Flow Surrounding Static Sub-Graphs:} We
observe that for most DL computations exhibiting control flow
dynamism, the dynamic control flow \emph{surrounds} tensor
computations. Consider the simple sequential RNN model implemented by
the \verb+@rnn+ function shown in Listing~\ref{code:rnn_relay}. Here,
we see that the sequential control flow surrounds an RNN cell on
lines~\ref{line:input_transform} and~\ref{line:state_transform}, which
is a static sub-graph of tensor computations with no intervening
control flow.

\noindent\textbf{Tensor-Dependent Control Flow:} Control flow
decisions often depend on the values of intermediate tensors in DL
computations. Examples of such models and computations include beam
search in machine translation, StackLSTMs~\cite{stacklstm},
Tree-to-Tree neural networks (T2TNN)~\cite{tree2tree}, models with
early exits~\cite{early_exit_cnn1, early_exit_cnn2, deebert,
  depth_adaptive_transformer} and Mixture-of-Experts~\cite{moe1, moe2,
  switch_trans}. Meanwhile, in models such as
TreeLSTM~\cite{treelstm}, DAG-RNN, sequential RNNs and their variants,
control flow only depends on the inputs and not on intermediate
tensors.

\noindent\textbf{Repetitive Control Flow:} We say that a model
exhibits repetitive control flow if it can be expressed as an
iterative or recursive computation. This includes iterative models
such as RNNs and their variants (LSTM and GRU~\cite{gru} for example)
and StackLSTMs, and recursive models such as TreeLSTM, Tree-to-Tree
neural networks and DAG-RNNs~\cite{dagrnn}. On the other hand,
Mixture-of-Experts and early exit models do not exhibit repetitive
control flow. Such models contain conditional execution in an
otherwise static feed-forward network. Repetitive control flow can
often also be nested. The GraphRNN model, for example, executes two
RNNs, one nested inside the other. Similarly, the DRNN model, which is
used for top-down recursive tree generation, involves iterative
generation of children for a given tree node.

The presence of recursive, as opposed to iterative control flow, can
often complicate static analysis as parallelism is more easily
exploited with the latter. We see in \S\ref{sec:tdc_fiber} how
exploiting parallelism across recursive calls at runtime, for example,
can require the multiple concurrent execution contexts, similar to the
fork-join parallelism paradigm~\cite{fork_join}.

\noindent\textbf{Control-Flow in Training and Inference:} We see, in
Table~\ref{table:dynamism}, that the computation for a lot of the
models involve dynamic control flow during both training as well as
inference. This is however, not the case for models with early exits,
where during training, we often wish to train all the exit branches
rather than evaluating one, as is the case during inference. Further,
search procedures such as beam search are often used only during
inference and hence the underlying model may not exhibit dynamism
during training (unless the model computation itself involves
dynamism, as in the case of RNN models, for example).

\noindent\textbf{Control Flow Parallelism:} Dynamic control flow can
lead to parallelism in a DL computation. Such a computation may
exhibit (1) \emph{Batch Parallelism} that exists across different
input instances in the mini-batch, and/or (2)
\emph{Instance Parallelism} which refers to the parallelism that arises
due to dynamic control flow dependences, such as recursive
parallelism. The amount of such parallelism differs widely across
computations. Recursive models, often (though not always) have
significant parallelism across different recursive
calls. Correspondingly, iterative computations may contain loops that
can be executed concurrently. An example is the call to the
\verb+@map+ function call in the RNN implementation in
Listing~\ref{code:rnn_relay}.

\subsection{Dynamic Batching}\label{sec:dyn_batch}
\Sys~builds upon dynamic batching~\cite{tffold, dynet2}, a prior
technique to perform auto-batching in the presence of dynamic control
flow. Given a mini-batch of input instances, dynamic batching involves
lazily executing the model computation for each input instance while
building dataflow graphs (DFGs) of tensor operators for each instance
in the background. The execution of these DFGs is triggered when the
value of a particular tensor is requested (when the model contains
tensor-dependent control flow, for example). During this execution,
the runtime can identify batching opportunities within the DFGs and
launch batched kernels appropriately.

%% file: src/overview.tex
\section{\Sys: Overview and API}\label{sec:overview}
Control flow dynamism necessitates reliance on potentially expensive
runtime analysis for auto-batching. In \Sys, we observe that
aggressive static analysis often provides sufficient information to
reduce the overheads of such analyses. Such analyses further enable us
to generate specialized and more efficient tensor kernels in an
end-to-end manner.

\begin{figure*}
  \centering
  \vspace{-2mm}
  \includegraphics[width=0.95\linewidth]{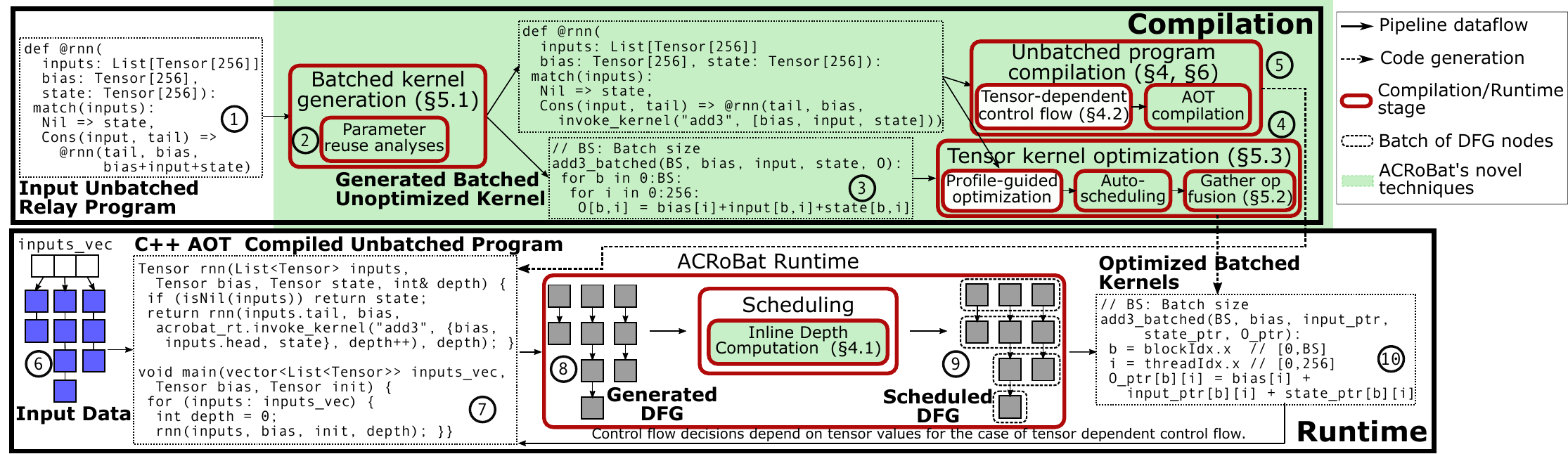}
  \vspace{-3mm}
  \caption{Overview of \Sys's
    workflow. Fig.~\ref{fig:ap_dynet_overview} in the appendix shows a
    corresponding overview of DyNet, a prior fully dynamic
    approach. Note how \Sys~performs significant novel analysis and
    code generation at compile-time to reduce runtime overheads.}
  \vspace{-4mm}
  \label{fig:overview}
\end{figure*}

\ifminted

\begin{listing}[h]
\begin{minted}[escapeinside=||,linenos,numbersep=4pt,frame=lines,fontsize=\tiny]{ocaml}
def @rnn(inps, state, bias, i_wt, h_wt) {
  match(inps) {
    Nil => Nil,
    Cons(inp, tail) => {
      let inp_linear = bias + nn.dense(inp, i_wt);                    |\label{line:input_transform}|
      let new_state = sigmoid(inp_linear + nn.dense(state, h_w));      |\label{line:state_transform}|
      Cons(new_state, @rnn(tail, new_state, bias, i_w, h_w))
    }}}

def @main(rnn_bias: Tensor[(1, 256)], rnn_i_wt: Tensor[(256, 256)],
          rnn_h_wt: Tensor[(256, 256)], rnn_init: Tensor[(1, 256)],
          c_wt: Tensor[(16, 512)], cbias: Tensor[(1, 16)],
          inps: List[Tensor[(1, 256)]]) {
  (* Recursive computation stage (program phase 1) *)
  let rnn_res =
    @rnn(inps, rnn_init, rnn_bias, rnn_i_wt, rnn_h_wt);
  (* Output transformations stage (program phase 2) *)
  @map(fn(p: Tensor[(1, 256)]) {
         nn.relu(cbias + nn.dense(p, c_wt))           |\label{line:output_op}|
       }, rnn_res) }
\end{minted}
\vspace{-5mm}
\caption{A simple RNN model expressed in a functional language (here,
  Relay~\cite{relay} is used for illustration) as an input to \Sys.}
\vspace{-4mm}
\label{code:rnn_relay}
\end{listing}

\else

\lstset
{ 
    language=C++,
    basicstyle=\footnotesize,
    numbers=left,
    stepnumber=1,
    showstringspaces=false,
    tabsize=1,
    breaklines=true,
    breakatwhitespace=false,
}

 \begin{lstlisting}[language=C++,basicstyle=\scriptsize, captionpos=b, caption={A simple RNN model expressed in Relay as an input
  to \Sys.}, label={code:rnn_relay}]
def @rnn(inputs, state, bias, i_wt, h_wt) {
  match(inputs) {
    Nil => Nil,
    Cons(input, tail) => {
      let input_linear = bias + nn.dense(input, i_wt);
      let new_state = sigmoid(input_linear + nn.dense(state, h_wt));
      Cons(new_state, @rnn(tail, new_state, bias, i_wt, h_wt))
    }
  }
}

def @main(rnn_bias: Tensor[(1, 256)], rnn_i_wt: Tensor[(256, 256)],
          rnn_h_wt: Tensor[(256, 256)], rnn_init: Tensor[(1, 256)],
          c_wt: Tensor[(16, 512)], cbias: Tensor[(1, 16)],
          inputs: List[Tensor[(1, 256)]]) {
  let _ = db.set_phase(0);
  let rnn_res =
    @rnn(inputs, rnn_init, rnn_bias, rnn_i_wt, rnn_h_wt);
  let _ = db.set_phase(1);
  @map(fn(p: Tensor[(1, 256)]) {
         nn.relu(cbias + nn.dense(p, c_wt))
       }, rnn_res)
}
\end{lstlisting}

\fi

We will now look at \Sys's compilation and execution workflows
(illustrated in Fig.~\ref{fig:overview}) that make use of the above
insights. \Sys~has been designed to take an unbatched DL computation
expressed in a simple Turing-complete functional language as an input.
This enables \Sys~users to easily express models with dynamic control
flow, such as the ones discussed in~\S\ref{sec:survey}. For example,
Listing~\ref{code:rnn_relay} illustrates a simple RNN model which
\Sys~can take as an input.

Given an input computation \circled{1}, compilation in \Sys~begins
with batched kernel generation \circled{2}. Here, \Sys~performs novel
static analysis (\S\ref{sec:param_reuse}) to identify data reuse
opportunities and accordingly generates batched kernels \circled{3}
implementing the tensor operators used in the input program. Further,
gather operator fusion (\S\ref{sec:gather_fuse}) enables us to
generate specialized kernels that minimize data movement. These
unoptimized kernels are then optimized by an auto-scheduler
\circled{4}. Once optimized, target code \circled{10} such as CUDA C++
can be generated for the batched kernels. Concurrently, the input
program is further optimized and compiled \circled{5} in an
ahead-of-time (AOT) fashion to generate C++ code \circled{7}. As part
of this compilation, \Sys~generates code to (1) enable low overhead
scheduling via our inline depth computation approach, and (2)
automatically enable concurrent execution in the presence of tensor
dependent control flow (\S\ref{sec:tdc_fiber}).

At runtime, \Sys~lazily executes the AOT compiled input program
\circled{7} on a mini-batch of inputs \circled{6}, and constructs DFGs
\circled{8}. The \Sys~runtime library will then schedule these DFGs
(using inline depth computation as mentioned above) \circled{9}, while
looking for batching opportunities. Then, it will invoke the optimized
batched kernels \circled{10} for each identified batch of DFG
nodes. If the input program exhibits tensor dependent control flow,
the execution cycles back to the AOT compiled program which will
execute further and create more DFGs.

We will now take a look at \Sys's hybrid optimizations in
\S\ref{sec:hybrid_opt} and its tensor kernel generation in
\S\ref{sec:kernels}.

%% file: src/hybrid_analysis.tex
\section{Hybrid Static+Dynamic Optimizations}\label{sec:hybrid_opt}
Dynamic control flow often precludes static program
transformations. Therefore, \Sys~takes a hybrid approach whereby it
exploits static program knowledge by either (1) providing hints to the
dynamic analysis (\S\ref{sec:inline_depth}), or (2) generating code
that affords the dynamic analysis greater freedom in exploiting
parallelism (\S\ref{sec:tdc_fiber}). Further, static analysis also
enables us to perform optimizations such as kernel fusion, which is
important for high performance (\S\ref{sec:opt_eval}). Below, we
provide more details regarding our hybrid analysis.

\subsection{Inline Depth Computation}\label{sec:inline_depth}
As past work~\cite{cortex} has noted, prior fully dynamic approaches
incur significant scheduling overheads. For instance, as we will show
in Table 5, DyNet's scheduling overheads dominate the time spent in
tensor computations for the TreeLSTM model. Instead, as described
below, \Sys~devises a scheme to perform scheduling as it constructs
the DFGs, thereby lowering scheduling overheads greatly
(\S\ref{sec:eval}).

A DFG scheduling algorithm has two goals:%
\begin{enumerate}[label=\textbf{G.\arabic*}, topsep=-1ex, leftmargin=2em, itemsep=-0.7ex]
\item \label{go:correct} \textbf{Correctness}: Scheduling tasks such
  that dependences between the tasks are respected.
\item \label{go:perf} \textbf{Performance}: Identifying and exploiting
  parallelism.
\end{enumerate}

Given a DFG(s), we can satisfy both these goals by executing DFG nodes
(each of which represents one tensor operator) in the increasing order
of their topological depth\footnote{If $\mathcal{P}(n)$ denotes all
the set of all producers of all the tensors a node $n$ consumes, then
its depth $d_n$ is given by $d_n = 1 + \max_{p \in \mathcal{P}(n)}
d_p$ if $\mathcal{P}(n) \ne \varnothing$ and $0$ otherwise.}, such
that nodes at the same depth are executed concurrently~\cite{dynet1,
  tffold}. We make the following two observations in order to compute
these depths during DFG construction:

\begin{enumerate}[label=\textbf{O.\arabic*}, topsep=-1ex, leftmargin=2em, itemsep=-0.7ex]
\item \label{ob:correct} The order in which the unbatched program
  invokes the tensor operators, i.e. the order in which nodes are
  added to the DFGs, is a valid dependency order.
\item \label{ob:perf} Information about instance parallelism (for
  example, recursive parallelism in the TreeLSTM model as seen in
  Table~\ref{table:dynamism}) is often available during compilation.
\end{enumerate}

\ifminted
\definecolor{lightyellow}{RGB}{255,255,150}
\setlength{\fboxsep}{2pt}
\begin{listing}[h]
\begin{minted}[escapeinside=||,linenos,numbersep=4pt,frame=lines,fontsize=\tiny]{c++}
List<Tensor> rnn(List<Tensor> inps, Tensor state, Tensor bias,
                 Tensor i_wt, Tensor h_wt, |\colorbox{lightyellow}{int& depth)}| {
  if (inps == ListNil()) return ListNil();
  auto inp_linear = AcrobatRT.InvokeKernel("bias_dense",
                        0, {bias, i_wt, inps.head});                 |\label{line:inp_linear_aot}|
  auto new_state = AcrobatRT.InvokeKernel("sigmoid_add_dense",
                     |\colorbox{lightyellow}{depth++}|, {inp_linear, h_wt, state});
  return ListCons(new_state, rnn(inps.tail, state, bias, i_wt,
				 h_wt, |\colorbox{lightyellow}{depth}|)); }

vector<Tensor> main(Tensor rnn_bias, Tensor rnn_i_wt,
		    Tensor rnn_h_wt, Tensor rnn_init, Tensor c_wt,
                    Tensor cbias, vector<List<Tensor>> inps_vec) {
  vector<Tensor> res;
  for (auto inps: inps_vec) {
    |\colorbox{lightyellow}{int depth = 0;}|
    /* Recursive computation stage (program phase 1) */
    auto rnn_res = rnn(inps, rnn_init, rnn_bias, rnn_i_wt,
                       rnn_h_wt, |\colorbox{lightyellow}{depth}|);
    /* Output transformations stage (program phase 2) */
    |\colorbox{lightyellow}{depth++;}|
    res.push_back(map([&](Tensor p) { AcrobatRT.InvokeKernel(
      "relu_bias_dense", |\colorbox{lightyellow}{depth}|, {cbias, c_wt, p}); }, rnn_res)); }         |\label{line:map_aot}|
  return res; }
\end{minted}
\vspace{-5mm}
\caption{AOT compiled output for the RNN model in
  Listing~\ref{code:rnn_relay}, with inline depth computation code
  highlighted.}
\label{code:rnn_aot}
\end{listing}

\else

\lstset
{ 
    language=C++,
    basicstyle=\footnotesize,
    numbers=left,
    stepnumber=1,
    showstringspaces=false,
    tabsize=1,
    breaklines=true,
    breakatwhitespace=false,
}

 \begin{lstlisting}[language=C++,basicstyle=\scriptsize, captionpos=b, caption={The output of the AOT compiler for the RNN model
  in Listing~\ref{code:rnn_relay}, with inline depth computation.}, label={code:rnn_aot}]
List<Tensor> rnn(List<Tensor> inputs, Tensor state, Tensor bias,
		 Tensor iweight, Tensor hweight, int& depth) {
  if (inputs == ListNil()) return ListNil();
  auto input_linear = AcrobatRT.InvokeKernel("bias_dense",
                        0, {bias, iweight, inputs.head});
  auto new_state = AcrobatRT.InvokeKernel("sigmoid_add_dense",
                     depth++, {input_linear, hweight, state});
  return ListCons(new_state, rnn(inputs.tail, state, bias,
				 iweight, hweight, depth));
}

vector<Tensor> main(Tensor rnn_bias, Tensor rnn_iweight,
		    Tensor rnn_hweight, Tensor rnn_init, Tensor cweight,
                    Tensor cbias, vector<List<Tensor>> inputs_vec) {
  vector<Tensor> res;
  for (auto inputs: inputs_vec) {
    int depth = 0;
    AcrobatRT.SetPhase(0);
    auto rnn_res = rnn(inputs, rnn_init, rnn_bias,
                       rnn_iweight, rnn_hweight, depth);
    AcrobatRT.SetPhase(1);
    depth++;
    res.push_back(map([&](Tensor p) { AcrobatRT.InvokeKernel(
      "relu_bias_dense", depth, {cbias, cweight, p}); }, rnn_res));
  }
  return res;
}
\end{lstlisting}

\fi

Based on these observations, we set the depth of an operator to be
equal to its position in the dependency ordering induced by the
execution of the unbatched program, thus meeting
goal~\ref{go:correct}. Then, we rely on observation~\ref{ob:perf}
above in order to discover and exploit opportunities for parallelism
by using the following techniques:

\addtolength{\columnsep}{-2pt}%
\begin{wrapfigure}{r}{0.28\columnwidth}
  \begin{center}
    \includegraphics[width=0.28\columnwidth]{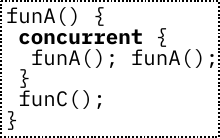}
  \end{center}
  \vspace{-5mm}
  \caption{Concurrent call annotation.}
  \label{fig:concurrent}
\end{wrapfigure}
\addtolength{\columnsep}{2pt}%

\noindent\textbf{Instance Parallelism:} We note that instance
parallelism often stems from recursion or the use of the functional
\verb+@map+ function on a list of independent items
(observation~\ref{ob:perf}). We ensure that such concurrent operators
are assigned the same depth during the execution of the unbatched
program. We rely on simple user annotations to obtain information
about recursive parallelism\footnote{Users can mark a set of function
calls as concurrent in the input code. Of the seven models we evaluate
in~\S\ref{sec:eval}, four required one such annotation each, while the
rest did not require any.}. Fig.~\ref{fig:concurrent} shows an example
where the two recursive calls to \verb+funA+ are annotated as
concurrent. Note also that past work
auto-parallelization~\cite{autopar1, autopar2} could potentially be
used in lieu of such annotations. Listing~\ref{code:rnn_aot} shows the
AOT compiled code generated for the RNN model in
Listing~\ref{code:rnn_relay}. We see, on line~\ref{line:map_aot}, how
all invocations of the \texttt{relu\_bias\_dense} kernel inside the
\verb+@map+ function are assigned the same depth.

\noindent\textbf{Combating Eagerness of Depth Scheduling:} As noted in
past work~\cite{dynet2}, a depth-based scheduling scheme, like the one
\Sys~uses, can often be too eager in executing tensor operators,
leading to a sub-optimal amount of exploited parallelism. Past work
has relied on \emph{agenda-based scheduling}~\cite{dynet2}, a more
expensive scheduling scheme, as an alternative to the depth-based
scheme to alleviate this problem. \Sys~instead relies on compile-time
analysis, as described below.

\begin{figure}
  \centering
  \includegraphics[width=0.8\columnwidth]{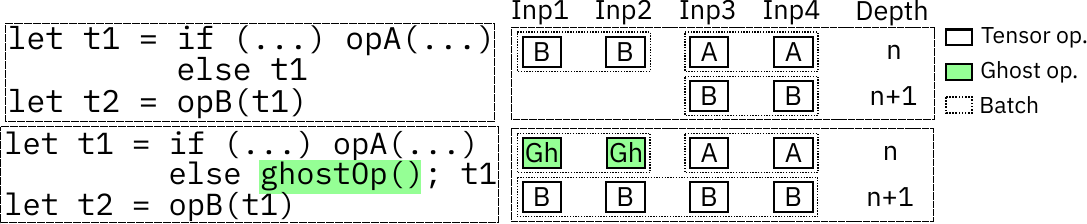}
  \vspace{-2mm}
  \caption{Ghost operators can enable better batching.}
  \vspace{-5mm}
  \label{fig:ghost_ops}
\end{figure}

\emph{Ghost Operations:} In the presence of conditional if statements,
eager batching leads to sub-optimal batching as illustrated in the
upper panes of Fig.~\ref{fig:ghost_ops}. We see that eager batching
leads to a sub-optimal batching schedule as the instances of operation
B for inputs Inp1 and Inp2 are batched eagerly and more importantly
separately from the instances of operation B for inputs Inp3 and
Inp4. In such situations, \Sys~can statically insert \emph{ghost
operations} to essentially delay the scheduling and execution of
certain operators, as shown in the lower panes of the figure. Note
that ghost operations merely affect \Sys's scheduling behavior and the
are ignored during tensor kernel execution.

\emph{Program Phases:} On the other hand, when repetitive (recursive
or iterative) control flow is present, we rely on \emph{program
phases}~\cite{phase1} to combat the aforementioned sub-optimality of
the scheduling. Given knowledge of such program phases, \Sys~waits to
schedule and execute operators in a phase until operators in all
previous phases have been scheduled and executed. We find that
considering individual semantic stages of the input DL computation as
individual phases is a good heuristic for dividing the computation
into phases. \Sys~also provides a way for users to override this
heuristic by manually annotate program phases, though in our
evaluation, we did not need such annotations. We provide more details
and explanations about program phases and ghost operations in
\S\ref{sec:ap_eager} of the appendix.

Further, \Sys~is also able to statically hoist operators, which we
describe in more detail in \S\ref{sec:ap_hoist} of the appendix. As an
example, in Listing~\ref{code:rnn_aot}, the invocation of the kernel
\texttt{bias\_dense} on line~\ref{line:inp_linear_aot} is assigned a
statically computed depth of 0, which during runtime, effectively
hoists the kernel invocation out of the recursion.

\subsection{Tensor Dependent Control Flow}\label{sec:tdc_fiber}
\begin{figure}
  \centering
  \includegraphics[width=0.9\columnwidth]{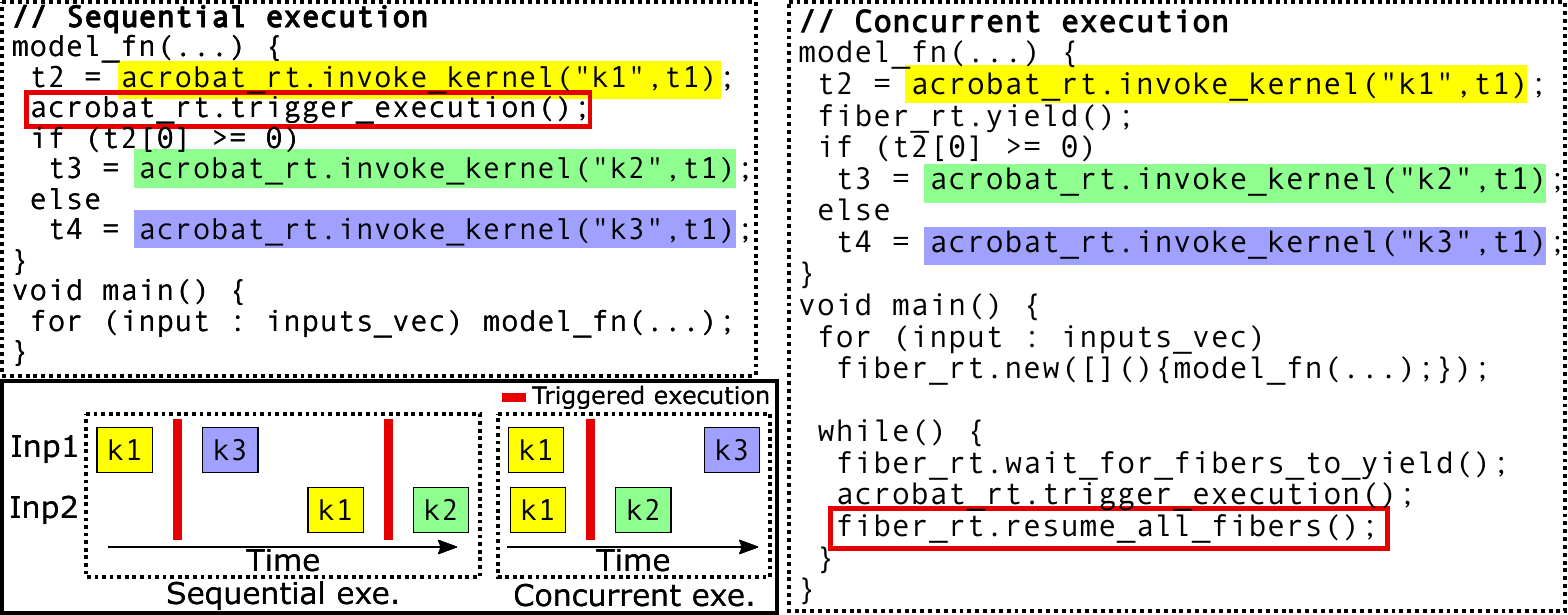}
  \vspace{-3mm}
  \caption{Concurrent execution of the unbatched program in the
    presence of tensor-dependent control flow.}
  \label{fig:tdc_fiber}
  \vspace{-3mm}
\end{figure}

\Sys~executes the unbatched program lazily to create DFGs for each
input instance in the batch. In the absence of tensor dependent
control flow, we can first execute the unbatched program for each
instance sequentially and trigger the batching and execution of all
the DFGs at once. In the presence of tensor dependent control flow,
however, such sequential execution would not allow us to exploit any
batch parallelism as we would be required to trigger the execution at
control flow decisions that depend on the value of intermediate
tensors. While prior work places the burden of restructuring input
computations to alleviate this issue on the user, \Sys~automatically
generates code to execute the unbatched program for each input
instance concurrently by using
\emph{fibers}\footnote{Fibers~\cite{fibers} allow multiple execution
stacks to be cooperatively scheduled on a single process.}. This way,
the unbatched programs can be executed for each instance to a point
where none can progress without triggering the evaluation of the
DFG. At this point, the evaluation can be performed, and the
concurrent executions resumed after as illustrated in
Fig.~\ref{fig:tdc_fiber}. Correspondingly, in order to exploit
instance parallelism in the presence of tensor dependent control flow,
\Sys~launches concurrent fibers, similar to the fork-join model of
parallelism~\cite{fork_join}. \Sys~thus combines the static knowledge
of parallelism with dynamic concurrent execution as part of its hybrid
analysis to effectively exploit parallelism in the presence of tensor
dependent control flow.

%% file: src/kernels.tex
\section{End-to-end Tensor Kernel Generation}\label{sec:kernels}
As we alluded to above, \Sys~enables end-to-end, uniform and automatic
tensor kernel code generation by avoiding the use of vendor
libraries. This allows \Sys~to support a larger set of operators
without additional compiler development effort. More details about
\Sys's tensor kernel generation are provided below.

\subsection{Exploiting Parameter Reuse}\label{sec:param_reuse}
Given the input unbatched computation, \Sys~needs to generate batched
kernels implementing the tensor operators used in the
computation. Generating these kernels is not straightforward because
some input tensors (often model parameters) might be shared across
calls to the operator. For example, across multiple calls to the
element-wise addition operator \texttt{add3} used in the input
computation \circled{1} in Fig.~\ref{fig:overview}, the \verb+bias+
argument will be shared (as it is a model parameter) and hence should
be reused across all values of the arguments \verb+input+ and
\verb+state+. This can be seen in the corresponding batched kernel
(\circled{3} and \circled{10}) in Fig.~\ref{fig:overview}.

A completely dynamic approach to auto-batching, such as the one used
in DyNet, is unable to accurately identify such parameter reuse, and
instead relies on heuristics, which can be brittle, leading to
sub-optimal performance (\S\ref{sec:perf_eval}). On the other hand,
\Sys~uses a 1-context sensitive\footnote{Context sensitivity is a
static analysis technique that allows the compiler to reason about a
function in the different contexts it may be called under leading to
increased analysis precision. For the DL computations we worked with,
we found that a 1-context sensitive analysis was sufficient. Deeper
contexts might be useful, however, for more complex computations.}
taint analysis to identify such shared arguments to tensor
operators. The use of static analysis here allows \Sys~to obtain
accurate knowledge about the parameter reuse patterns.

Beyond the analysis described above, \Sys~further explores
opportunities for data reuse by employing code duplication and
horizontal fusion as described in \S\ref{sec:ap_reuse}.

\subsection{Fusing Memory Gather Operations}\label{sec:gather_fuse}
As \Sys~identifies batching opportunities across the DFGs dynamically,
the input tensors to all DFG nodes in a batch may not be laid out
contiguously in the accelerator's memory. In this scenario, prior work
performs a memory gather before operating on the tensors (by invoking
vendor library kernels), leading to significant data movement
(\S\ref{sec:opt_eval}). Instead, \Sys~generates specialized batched
kernels to directly operate on tensors scattered in memory, in effect
fusing the expensive gather operation with the batched kernel. The
generated batched kernel \circled{10} in Fig.~\ref{fig:overview}
illustrates this. This fusion can lead to a significant performance
improvement as seen in \S\ref{sec:eval}.

%% file: src/implementation.tex
\section{Implementation Details}\label{sec:impl}
Our prototype of \Sys~is built upon TVM~\cite{tvm} v0.9.dev0, a DL
framework and a tensor compiler. It thus accepts as input computations
expressed in Relay. Our prototype, \Sys~also performs the grain size
coarsening optimization~\cite{jit_batch, cavs, cortex, batchmaker,
  ebatch}, which is discussed more in \S\ref{sec:ap_coarsen} of the
appendix.

\begin{table}
  \centering
  \scriptsize
  \vspace{-4mm}
  \caption{Models and datasets used in the evaluation.}
  \addtolength{\tabcolsep}{-3pt}
\resizebox{0.99\columnwidth}{!}{%
  \begin{tabular}{L{1.3cm}L{3.8cm}L{3.05cm}}
    \toprule
    Model       & Description                             & Dataset \\ \midrule
    TreeLSTM    & TreeLSTM                          & Stanford sentiment treebank~\cite{sst}         \\ \hline
    MV-RNN      & MV-RNN                            & Stanford sentiment treebank                    \\ \hline
    BiRNN       & Bidirectional RNNs         & XNLI~\cite{xnli}                               \\ \hline
    NestedRNN   & An RNN loop nested inside a GRU loop    & GRU/RNN loops iterate for a random number of iterations in [20, 40].  \\ \hline
    DRNN        & Doubly recurrent neural networks for top-down tree generation & Randomly generated tensors. \\ \hline
    Berxit      & Early exit for BERT inference~\cite{berxit}. All layers share weights.   & Sequence length 128.\\ \hline
    StackRNN    & StackLSTM parser with LSTM cells replaced by RNN cells.   & XNLI   \\
    \bottomrule
  \end{tabular}%
  }
  \addtolength{\tabcolsep}{3pt}
  \label{table:models}
  \vspace{-2mm}
\end{table}

As demonstrated in~\S\ref{sec:ap_aot_eval}, we find that using an
interpreted virtual machine (VM) for executing the unbatched programs
can incur significant VM overheads in the presence of control flow
dynamism. Therefore, \Sys~compiles the input computation to C++ in an
AOT fashion (as discussed in the appendix in
\S\ref{sec:ap_impl}). Further, as TVM does not support training, we
evaluate \Sys~for (batched) inference of DL computations. Other
implementation details, including those on \Sys's use of TVM's
auto-scheduler, can be found in the appendix in \S\ref{sec:ap_impl}.

%% file: src/evaluation.tex
\section{Evaluation}\label{sec:eval}
We now evaluate \Sys~against Cortex and DyNet on an Nvidia GPU. Cortex
and DyNet are both state-of-the-art auto-batching frameworks for DL
computations exhibiting recursive and general unrestricted control
flow respectively. They have been shown to be faster than generic
frameworks like PyTorch and TensorFlow~\cite{dynet1, dynet2,
  cortex}. We also compare \Sys's performance with that of PyTorch
(\S\ref{sec:pyt_eval}).

\subsection{Experimental Setup}
\noindent\textbf{Models:} We use the models listed in
Table~\ref{table:models} for the evaluation. For each model, we look
at two model sizes---small and large. For the MV-RNN model, we use
hidden sizes 64 and 128 for the small and large model sizes, while for
the Berxit model, the small model uses the same hyper-parameters as
the BERT\textsubscript{BASE} model~\cite{bert}, while the large model
uses the same hyper-parameters as the BERT\textsubscript{LARGE}
model~\cite{bert}, except that we use 18 layers instead of 24 in this
case. For the remaining models, the small and the large model sizes
use hidden sizes of 256 and 512 respectively.

\noindent\textbf{Experimental Environment:} We run our experiments on
a Linux workstation with an AMD Ryzen Threadripper 3970X CPU (64
logical cores with 2-way hyperthreading) and an Nvidia RTX 3070
GPU. The machine runs Ubuntu 20.04, CUDA 11.1 and cuDNN 8.0.5. We
compare against DyNet's commit 3e1b48c7 (March~2022) which uses the
Eigen library (v3.3.90).

\subsection{Benefits of AOT Compilation}\label{sec:ap_aot_eval}
We first look at the benefits of AOT compilation
(\S\ref{sec:impl}). The performance of the TreeLSTM, MV-RNN and BiRNN
models\footnote{\Sys's prototype implementation does not currently
support the execution of the remaining models in
Table~\ref{table:models} using the Relay VM.} when executed using the
Relay VM and \Sys's AOT compiler (with the grain size coarsening,
gather operator fusion and program phase optimizations turned on) is
shown in Table~\ref{table:ap_aot_eval}. We see that overheads
significantly slow down the execution (by up to 13.45$\times$) as
compared to the AOT compiled native code for these models. Therefore,
for the rest of this section, we evaluate \Sys's performance with AOT
compilation turned on.

\begin{table}[t]
  \centering
  \scriptsize
  \caption{Relay VM vs.~\Sys's AOT compilation: Inference latencies in
    $ms$.}
  \begin{tabular}{cc cc cc cc}
    \toprule
    \multirow{2}{6.5mm}{\centering Hidden Size} & \multirow{2}{6.5mm}{\centering Batch Size} & \multicolumn{2}{c}{TreeLSTM} &  \multicolumn{2}{c}{MV-RNN}  & \multicolumn{2}{c}{BiRNN} \\
    \cmidrule(lr){3-4} \cmidrule(lr){5-6} \cmidrule(lr){7-8}
    &  & VM & AOT  & VM & AOT  & VM & AOT  \\ \midrule
small & 8 & 30.68 & 2.66 & 4.0 & 0.55 & 29.88 & 2.23 \\
small & 64 & 28.94 & 9.47 & 3.91 & 1.63 & 28.88 & 5.47 \\
large & 8 & 31.64 & 3.85 & 4.34 & 1.06 & 32.04 & 4.82 \\
large & 64 & 29.49 & 15.9 & 4.36 & 4.6 & 30.43 & 13.72 \\
    \bottomrule
  \end{tabular}
  \label{table:ap_aot_eval}
\end{table}

\subsection{Overall Performance}\label{sec:perf_eval}
In this section, we compare \Sys's performance with that of PyTorch,
DyNet and Cortex.

\textbf{Performance Comparison with PyTorch}\\%
\begin{figure}
  \centering
  \includegraphics[width=0.93\columnwidth]{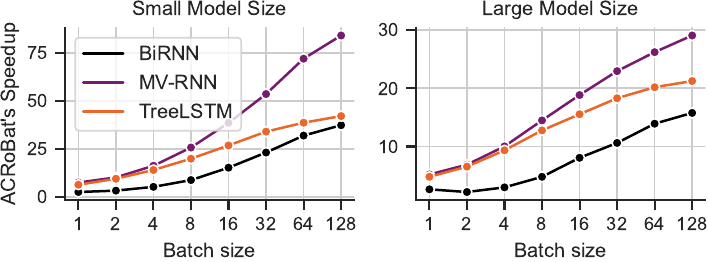}
  \caption{Speedups obtained over PyTorch for the TreeLSTM, MV-RNN and
    BiRNN models.}
  \label{fig:pytorch_eval}
\end{figure}%
Fig.~\ref{fig:pytorch_eval} compares \Sys's performance with that of
PyTorch (v1.9.0a0+gitf096245) for the TreeLSTM, MV-RNN and BiRNN
models\footnote{We use TorchScript only for the BiRNN model as it does
not currently support recursive data
types~\cite{torchscript_recursive}, such as the parse trees the
TreeLSTM and MV-RNN models operate on.}. PyTorch does not perform
auto-batching and is therefore unable to exploit any available
instance or batch parallelism in the evaluated computations. Further,
\Sys's kernel fusion and other static optimizations also increase its
performance relative to PyTorch. The speedups are higher for the small
model size as compared to the larger model sizes because the relative
importance of exploiting instance and batch parallelism is lower for
the large model size due to the increased parallelism in individual
tensor operators. \Sys's relatively worse performance on the BiRNN
model as compared to the other two can be attributed to the absence of
instance parallelism in BiRNN leading to a lower amount of parallelism
that \Sys~can exploit. Similarly, due to TreeLSTM exhibiting a higher
amount of static and tensor parallelism as compared to MV-RNN, the
relative importance of exploiting instance and batch parallelism is
lower, leading to performance lower than that of MV-RNN.%

\textbf{Performance Comparison with DyNet} \\
We now compare \Sys's performance with that of DyNet. As mentioned in
\S\ref{sec:impl}, TVM does not support the training of DL
models. Therefore, due to lack of access to trained model parameters,
we use pseudo-randomness to emulate tensor dependent control flow in
the NestedRNN, DRNN, Berxit and StackRNN models as part of our
evaluation. We ensure that the pseudo-randomness is uniform across the
\Sys~and DyNet implementations by using pre-determined random seeds
for a fair comparison. An exception is the DRNN model when inline
depth computation is performed. In this case, \Sys~exploits DRNN's
recursive instance parallelism using fibers (\S\ref{sec:tdc_fiber})
leading to a change in the random control flow decisions taken. We
account for this by presenting the mean execution time across 50
different random seeds.

\begin{table*}[t]
  \centering
  \scriptsize
  \vspace{-4mm}
  \caption{DyNet vs.~\Sys: Inference latencies (DyNet/\Sys) in $ms$
    and speedups. The DyNet implementation of the Berxit model was
    killed due to out-of-memory errors for a batch size of 64.}
  \addtolength{\tabcolsep}{-2pt}
\resizebox{0.99\textwidth}{!}{%
  \begin{tabular}{cc cc cc cc cc cc cc cc}
    \toprule
    \multirow{2}{6.5mm}{\centering Hidden Size} & \multirow{2}{6.5mm}{\centering Batch Size} & \multicolumn{2}{c}{TreeLSTM} &  \multicolumn{2}{c}{MV-RNN}  & \multicolumn{2}{c}{BiRNN}  & \multicolumn{2}{c}{NestedRNN}  & \multicolumn{2}{c}{DRNN}  & \multicolumn{2}{c}{Berxit}  & \multicolumn{2}{c}{StackRNN} \\
    \cmidrule(lr){3-4} \cmidrule(lr){5-6} \cmidrule(lr){7-8} \cmidrule(lr){9-10} \cmidrule(lr){11-12} \cmidrule(lr){13-14} \cmidrule(lr){15-16}
     &  & Time & Speedup  & Time & Speedup  & Time & Speedup  & Time & Speedup  & Time & Speedup  & Time & Speedup  & Time & Speedup \\ \midrule
small & 8 & 4.31/\textbf{1.48} & 2.93 & 2.11/\textbf{0.54} & 3.96 & 3.13/\textbf{2.16} & 1.45 & \textbf{29.38}/31.01 & 0.95 & 6.7/\textbf{1.74} & 3.87 & 63.54/\textbf{38.49} & 1.66 & 47.78/\textbf{22.69} & 2.11 \\
small & 64 & 26.18/\textbf{5.81} & 4.51 & 12.45/\textbf{1.48} & 8.47 & 12.04/\textbf{4.86} & 2.49 & 84.55/\textbf{65.73} & 1.29 & 25.3/\textbf{5.24} & 4.84 & -/204.54 & - & 213.98/\textbf{39.06} & 5.48 \\
large & 8 & 4.58/\textbf{2.4} & 1.92 & 2.27/\textbf{1.04} & 2.19 & \textbf{3.95}/4.43 & 0.9 & 46.03/\textbf{35.61} & 1.3 & 8.44/\textbf{2.45} & 3.45 & 113.18/\textbf{64.49} & 1.76 & 64.67/\textbf{43.75} & 1.48 \\
large & 64 & 26.53/\textbf{11.44} & 2.33 & 13.89/\textbf{4.46} & 3.13 & \textbf{12.11}/13.11 & 0.93 & \textbf{94.97}/100.17 & 0.95 & 26.5/\textbf{9.99} & 2.66 & -/335.3 & - & 230.74/\textbf{86.82} & 2.66 \\
    \bottomrule
  \end{tabular}%
  }
  \addtolength{\tabcolsep}{2pt}
  \label{table:dynet_eval}
  \vspace{-5mm}
\end{table*}%

The execution latencies for DyNet and \Sys~are shown in
Table~\ref{table:dynet_eval}\footnote{We consider the best of the two
scheduling schemes DyNet implements~\cite{dynet2} for each model
configuration.}. \Sys~performs better than DyNet in most cases due to
a number of reasons. Table~\ref{table:profile_table} lists the time
spent by the frameworks for different runtime activities for the
TreeLSTM model. We see that \Sys's optimizations such as static kernel
fusion and grain size coarsening reduce the number of tensor kernels
invoked, thereby significantly reducing DFG construction and
scheduling overheads. Further, inline depth computation allows \Sys~to
exploit available parallelism with lower overheads. Optimizations such
as static kernel fusion and gather operator fusion enable \Sys~to
launch fewer GPU kernels, further reducing the time spent in the CUDA
API. We look at the benefits of each of \Sys's optimizations in more
detail in~\S\ref{sec:opt_eval}.

While, overall, \Sys~performs 2.3$\times$ better than DyNet across all
model configurations, DyNet performs slightly better than \Sys~for
some configurations of the BiRNN and NestedRNN models. For the former,
Table~\ref{table:profile_table} shows that while \Sys~incurs lower
runtime overheads for DFG construction, scheduling and memory
transfer, it spends a higher amount of time in kernel execution
compared to DyNet. We believe that better tensor kernel optimizations
can help reduce this performance gap.

\begin{table}
  \centering \scriptsize
  \vspace{-2mm}
  \begin{threeparttable}[b]
  \caption{Time spent ($ms$) in various activities\tnote{1} for DyNet
    and \Sys~for batch size 64.}
  \addtolength{\tabcolsep}{2pt}
  \addtolength\extrarowheight{-0.5pt}
  \begin{tabular}{c cc cc}
    \toprule
    \multirow{2}{6.5mm}{\centering Activity} & \multicolumn{2}{c}{TreeLSTM, small} &  \multicolumn{2}{c}{BiRNN, large} \\
    \cmidrule(lr){2-3} \cmidrule(lr){4-5}
                               & DyNet    & \Sys     & DyNet     & \Sys     \\
    \midrule
    DFG construction           & 8.8      & 1.5      & 4.5       & 1.0      \\
    Scheduling                 & 9.7      & 0.4      & 3.3       & 0.4      \\
    Memory copy time           & 3.1      & 0.1      & 2.3       & 0.2      \\
    GPU kernel time\tnote{2}   & 6.1      & 4.0      & 6.6       & 11.2     \\
    \#Kernel calls             & 1653     & 183      & 580       & 380      \\
    CUDA API time\tnote{3}     & 16.5     & 3.9      & 12.0      & 11.1     \\
    \bottomrule
  \end{tabular}
\addtolength\extrarowheight{0.5pt}
  \addtolength{\tabcolsep}{-2pt}
  \begin{tablenotes}
  \item [1] The timings reported correspond to multiple runs, and were
    obtained using manual instrumentation and profiling using Nvidia
    Nsight Systems. Due to profiling overheads, the execution times
    may not match the ones in Table~\ref{table:dynet_eval}.
  \item [2] Includes memory copy kernels.
  \item [3] Includes calls \verb+cudaMemcpy+, \verb+cudaMemcpyAsync+
    and all kernels.
  \end{tablenotes}
  \label{table:profile_table}
  \end{threeparttable}%
  \vspace{-7mm}
\end{table}%

Beyond the above reasons, \Sys~performs better on specific benchmarks
for the reasons discussed next.

\noindent\textbf{Accurate parameter reuse inference and automated
  batched kernel generation:} As mentioned in \S\ref{sec:param_reuse},
\Sys's use of static analysis for inferring parameter reuse allows it
to have accurate knowledge to statically generate the appropriate
batched kernels. On the other hand, DyNet's heuristic-based approach
is unable to batch instances of certain operators, forcing sequential
unbatched execution which leads to low performance. For instance,
DyNet heuristically batches multiple instances of the matrix
multiplication operator only when the first argument of all the
instances is the same tensor. This usually works as the first argument
is often a model parameter, usually as part of a linear
transformation. Our DyNet implementation of the MV-RNN model, however,
multiplies two intermediate tensor activations together, as a result
of which DyNet is unable to batch instances of this operator, forcing
sequential unbatched execution. When we modify DyNet's heuristic for
matrix multiplication, its performance improves significantly as shown
in Table~\ref{table:ap_dynet_pp}.

Further, as described in \S\ref{sec:kernels}, \Sys's end-to-end kernel
generation leads to a broader coverage over tensor operators for which
batching is supported as compared to approaches such as DyNet which
rely on vendor libraries. As a result, DyNet does not support batching
for certain operators, again leading to sequential execution and low
performance. Specifically, DyNet does not support batched execution
for the argmax operator, which the StackRNN model uses in order to
determine the next parser action in every iteration based on the
result of the embedded RNN cell. Similarly, the element-wise
multiplication operator, used in the DRNN model, is executed in an
unbatched manner when broadcasting needs to be performed. On the other
hand, \Sys~automatically generates optimized batched implementations
of these tensor operators. We also find that DyNet is unable to batch
calls to the operator that constructs constant tensors. We use this
operator to initialize the hidden states of tree leaves in the
TreeLSTM model. \Sys, on the other hand, statically recognizes that a
constant tensor can be reused and thereby only creates the tensor
once. The performance of the TreeLSTM model improves when we exploit
this reuse manually in DyNet, as Table~\ref{table:ap_dynet_pp} shows.

\begin{table}[t]
  \centering
  \scriptsize
  \caption{Model execution times in $ms$ after the improvements
    described in \S\ref{sec:perf_eval} were made for the TreeLSTM,
    MV-RNN and DRNN models. DN, DN++ and AB stand for DyNet, DyNet
    with improvements and \Sys~respectively.}
  \addtolength{\tabcolsep}{-3pt}
\resizebox{0.99\columnwidth}{!}{%
  \begin{tabular}{cc ccc ccc ccc}
    \toprule
    \multirow{2}{6.5mm}{\centering Model Size} & \multirow{2}{6.5mm}{\centering Batch Size} & \multicolumn{3}{c}{TreeLSTM} &  \multicolumn{3}{c}{MV-RNN}  & \multicolumn{3}{c}{DRNN} \\
    \cmidrule(lr){3-5} \cmidrule(lr){6-8} \cmidrule(lr){9-11}
     &  & DN & DN++ & AB & DN & DN++ & AB & DN & DN++ & AB \\ \midrule
small & 8 & 4.31 & 3.8 & 1.48 & 2.11 & 1.05 & 0.54 & 6.7 & 3.29 & 1.74 \\
small & 64 & 26.18 & 22.69 & 5.81 & 12.45 & 3.15 & 1.48 & 25.3 & 18.51 & 5.24 \\
large & 8 & 4.58 & 4.14 & 2.4 & 2.27 & 1.83 & 1.04 & 8.44 & 3.82 & 2.45 \\
large & 64 & 26.53 & 24.09 & 11.44 & 13.89 & 10.47 & 4.46 & 26.5 & 18.86 & 9.99 \\
    \bottomrule
  \end{tabular}%
}
  \addtolength{\tabcolsep}{3pt}
  \label{table:ap_dynet_pp}
\end{table}%

\noindent\textbf{Automated code generation for handling tensor
  dependent control flow:} The DRNN model constructs a tree from an
input vector representation in a top-down recursive manner. It
exhibits both tensor-dependent control flow as well as instance
parallelism (multiple subtrees can be generated concurrently). We saw
how \Sys~can automatically exploit instance parallelism in the
presence of tensor-dependent control flow with the use of fibers in
\S\ref{sec:tdc_fiber}. On the other hand, DyNet is unable to exploit
this parallelism and therefore \Sys's performance on this model is
significantly better than that of DyNet. Table~\ref{table:ap_dynet_pp}
also shows the performance improvement obtained in DyNet for the DRNN
model when the instance parallelism exhibited by the model computation
is manually exploited as detailed above.

\textbf{Performance Comparison with Cortex} \\
\begin{table}[t]
  \centering \scriptsize
  \caption{Cortex vs.~\Sys: Inference latencies in $ms$. Note that
    unlike \Sys, Cortex is limited to recursive computations, and does
    not support the other models in Table~\ref{table:models}. Further,
    Cortex places a high development burden on its users by relying on
    manual kernel optimization.}

  \addtolength{\tabcolsep}{-3pt}
  \resizebox{0.99\columnwidth}{!}{%
  \begin{tabular}{cc cc cc cc}
    \toprule
    \multirow{2}{6.5mm}{\centering Hidden Size} & \multirow{2}{6.5mm}{\centering Batch Size} & \multicolumn{2}{c}{TreeLSTM} &  \multicolumn{2}{c}{MV-RNN}  & \multicolumn{2}{c}{BiRNN} \\
    \cmidrule(lr){3-4} \cmidrule(lr){5-6} \cmidrule(lr){7-8}
    &  & Cortex & \Sys  & Cortex & \Sys  & Cortex & \Sys  \\ \midrule
small & 8 & \textbf{0.79} & 1.48 & 1.14 & \textbf{0.54} & \textbf{1.28} & 2.16 \\
small & 64 & \textbf{3.62} & 5.81 & 6.92 & \textbf{1.48} & \textbf{3.48} & 4.86 \\
large & 8 & \textbf{1.84} & 2.4 & 5.3 & \textbf{1.04} & \textbf{2.47} & 4.43 \\
large & 64 & \textbf{10.23} & 11.44 & 41.15 & \textbf{4.46} & \textbf{10.74} & 13.11 \\
    \bottomrule
  \end{tabular}%
}
  \addtolength{\tabcolsep}{3pt}
  \label{table:cortex_eval}
  \vspace{-6mm}
\end{table}%
Table~\ref{table:cortex_eval} compares the performance of \Sys~with
that of Cortex for the TreeLSTM, MV-RNN and the BiRNN models. Note
that this is not an apples-to-apples comparison because, Cortex, being
specialized for recursive computations, does not support general
control flow (as is present in the other models in
Table~\ref{table:models}) unlike \Sys~as mentioned in
Table~\ref{table:comp}. Further, Cortex places a high development
burden on users who are required to manually optimize and tune their
models for specific hardware, unlike \Sys's automatic kernel
generation\footnote{For example, implementing the MV-RNN model in
Cortex requires 325 LoC in Python, as compared to the 79 LoC of Relay
and 108 LoC of Python in \Sys.}. Similarly, while \Sys~can
automatically hoist the input linear transformations out of the
recursive computation in the TreeLSTM and BiRNN models (as described
in \S\ref{sec:ap_hoist}), they need to be manually hoisted and
offloaded to cuBLAS in the case of Cortex.

\begin{figure*}
  \centering
  \vspace{-1mm}
  \includegraphics[width=0.97\linewidth]{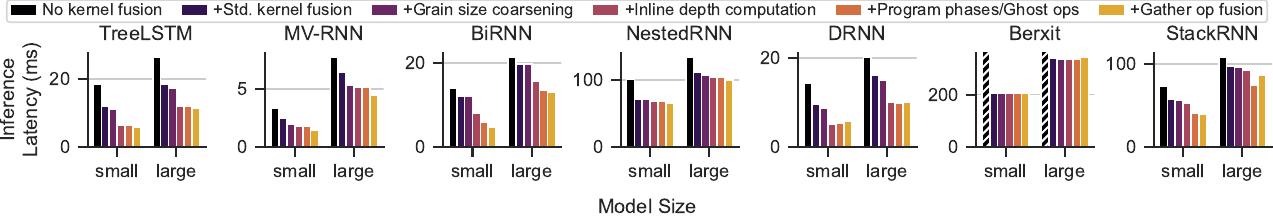}
  \vspace{-3mm}
  \caption{Benefits of different optimizations. The unfused executions
    of Berxit were killed due to out-of-memory errors.}
  \vspace{-5mm}
  \label{fig:opts_eval}
\end{figure*}

Being highly specialized for recursive computations, Cortex is able to
exploit aggressive kernel fusion, model persistence and incur low
kernel call overheads, thus performing up to 1.87$\times$ better than
\Sys~for the TreeLSTM and BiRNN models. However, note that Cortex
performs much worse than \Sys~on the MV-RNN model. This is because
Cortex's restrictive API necessitates additional copies of the
embedding vectors for the leaves of the input parse trees, which
\Sys~can avoid due to its more flexible interface. Overall,
\Sys~delivers performance comparable to that of Cortex, while
supporting a much wider range of DL computations with much lesser
developer effort.

\subsection{Benefits of Optimizations}\label{sec:opt_eval}
We now evaluate the relative benefits of the different optimizations
\Sys~performs. Fig.~\ref{fig:opts_eval} shows the execution times for
the models in Table~\ref{table:models} (at a batch size of 64) as we
progressively perform optimizations. Standard kernel fusion
(i.e.~kernel fusion not including gather operator fusion as discussed
in \S\ref{sec:gather_fuse}) provides significant benefits for all
models\footnote{The kernels used in the implementations with and
without standard kernel fusion were auto-scheduled for the same number
of auto-scheduler iterations.}. Grain size coarsening and inline depth
computation, both of which reduce scheduling overheads, are most
beneficial for models with a relatively high amount of control flow
such as TreeLSTM and MV-RNN. Further, in the case of the DRNN model,
inline depth computation also enables \Sys~to exploit the instance
parallelism inherent in the computation (\S\ref{sec:tdc_fiber})
leading to lower execution time. The BiRNN model involves per-token
output linear operators as in token classification. Here, program
phases allow \Sys~to batch all these operators together as described
in \S\ref{sec:inline_depth}. The StackRNN model executes different
tensor operators depending on the current parser action, which
involves a conditional statement. Ghost operators therefore enable
more optimal exploitation of parallelism leading to better
performance.

Gather operator fusion is advantageous for some benchmarks and but not
others. Such fusion leads to indirect memory accesses which can cause
a slowdown in the kernel execution. While \Sys~does hoist such loads
out of loops when appropriate, this is not always possible depending
on the schedule generated by the auto-scheduler. Further, gather
operator fusion leads to a slowdown mostly in models with iterative
execution and little instance parallelism. As in DyNet, when gather
operator fusion is turned off, \Sys~perform the explicit memory gather
only when the input tensors are not already contiguous in memory. This
is more likely to the case in such iterative models, thus blunting the
advantages of gather operator fusion. Also, in models such as Berxit,
the relatively high tensor computation cost of a coarsened static
block further reduces any benefits gather operator fusion might
provide.

Overall, models with a relatively lower amount of control flow or a
higher amount of tensor computations such as Berxit or NestedRNN or
models with the large size benefit less from optimizations that reduce
scheduling overheads.

%% file: src/related_work.tex
\section{Related Work}\label{sec:related}
\noindent\textbf{Auto-Batching for Dynamic Control Flow:} There has
been significant work on auto-batching techniques for dynamic
computations. Beyond dynamic batching (which is used in various forms
in DyNet, TensorFlow Fold, Cavs, Cortex and in
ByteTransformer~\cite{bytetransformer} specifically for transformer
models), static program transformations~\cite{matchbox, tf_pfor1,
  tf_pfor2, jax, radul_nuts} have also been explored for
auto-batching. Such techniques are often unable to fully exploit all
the available parallelism in the program as noted
in~\cite{radul_nuts}. \Sys~builds on these past techniques and
effectively uses both static as well as dynamic analysis thus
achieving lower runtime overheads while exploiting all the available
parallelism. Online batching approaches for low latency RNN inference
such as BatchMaker~\cite{batchmaker} and E-BATCH~\cite{ebatch} are
complementary to \Sys. \cite{dyn_batch_backprop} proposes improvements
to the dynamic batching technique for back propagation, while
ED-Batch~\cite{ed-batch} proposes efficient approaches to scheduling
and memory planning for the dynamic batching. These can be further
improved with \Sys's hybrid optimizations. Further, while grain size
coarsening has been explored in past work, we use it statically in the
context of general purpose auto-batching framework.




\noindent\textbf{Optimizing Dynamic DL Computations:} Beyond
auto-batching, there is a large body of work on optimizing the
execution of dynamic DL computations. Past work~\cite{janus, terra,
  lazy_tensor} has explored the lazy creation of DFGs that can be
optimized to accelerate dynamic models. There has also been
work~\cite{acs, grape} at better scheduling and low-overhead execution
tensor kernels to optimize for the dynamic execution patterns of
dynamic DL computations. Further SoD\textsuperscript{2}~\cite{sod2}
develops techniques for optimizing dynamic computations including
those with dynamic shapes. These techniques, which do not perform
batching, are complementary to \Sys's techniques. While \Sys~builds
upon TVM, our techniques can be implemented in other commonly used
compiler frameworks with expressive representations~\cite{torchscript,
  mlir} in a straightforward manner.

The gather operator fusion optimization is similar to the gather and
scatter fusion~\cite{cutlass_gather_scatter} performed for sparse GEMM
in the CUTLASS library though we perform this optimization
automatically as part of compilation. As mentioned in
\S\ref{sec:kernel_opt}, \Sys~borrows some techniques from DietCode for
efficient code generation. DietCode's techniques are complementary to
ours and it can be fully integrated into \Sys~for better kernel
performance.


\noindent\textbf{Traditional Compiler Techniques:} \Sys~uses
compilation techniques for programs written in general-purpose
languages. These include context-sensitivity~\cite{dragon_book}, taint
analysis which is extensively used for security
purposes~\cite{taint_analysis_security1, taint_analysis_security2},
profile-guided optimization~\cite{pgo_opt1, pgo_opt2} (as discussed
in~\S\ref{sec:kernel_opt} of the appendix) and program phases, which
have been used to adaptively optimize systems for different parts of a
program for optimal performance~\cite{phase_opt1, phase_opt2}. \Sys's
inline depth computation and DFG scheduling more generally are similar
to work on static and dynamic instruction scheduling for pipelined and
superscalar processors~\cite{dynamic_inst_sched1, dynamic_inst_sched2,
  static_inst_sched2, static_inst_sched1}. However, \Sys~applies these
techniques in the context of a DL framework.

%% file: src/conclusion.tex

\section{Conclusion}\label{sec:conclusion}
This paper presents \Sys, a compiler and runtime framework that
performs auto-batching of dynamic DL computations. \Sys~employs hybrid
static+dynamic analysis to enable effective batching with low runtime
overheads, and end-to-end code generation to generate highly optimized
tensor kernels for efficient execution. While we evaluated these
techniques only for the case of batched inference, we believe that
they also apply to DL training. In the context of the rising
importance of dynamism in DL computations, we believe that \Sys~is an
important step towards more collaborative relationships between
various components of a DL framework such as the tensor compiler, the
high-level language compiler and the runtime.

%% file: src/acks.tex
\section*{Acknowledgments}
This work was supported in part by grants from the National Science
Foundation (award CNS-2211882), Oracle, IBM, Qualcomm, DARPA (Real
Time Machine Learning, or RTML project) and by the Parallel Data Lab
(PDL) Consortium (Amazon, Facebook, Google, Hewlett-Packard
Enterprise, Hitachi, IBM, Intel, Microsoft, NetApp, Oracle, Pure
Storage, Salesforce, Samsung, Seagate, TwoSigma and Western
Digital). We would like to thank Saman Amarasinghe, Dominic Chen,
Siyuan Chen, Stephen Chou, Chris Fallin, Graham Neubig, Olatunji
Ruwase and the Catalyst Research Group at Carnegie Mellon University
for their valuable suggestions and feedback on our work.

%% file: src/appendix.tex
\begin{figure*}
  \centering
  \includegraphics[width=0.95\linewidth]{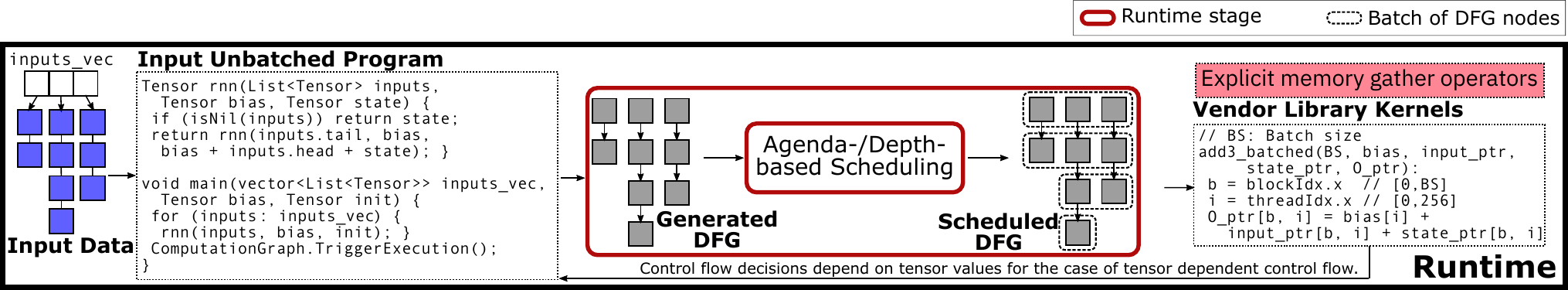}
  \vspace{-3mm}
  \caption{Overview of DyNet's runtime pipeline. Note the lack of any
    static, or compile-time analyses as well as how DyNet relies on
    explicit memory gather oprerations, leading to high data movement
    costs as we show in~\S\ref{sec:eval}.}
  \label{fig:ap_dynet_overview}
\end{figure*}

\section{More Details on Hybrid Static+Dynamic Optimizations}
\subsection{Operator Hoisting}
\label{sec:ap_hoist}
Given a recursive computation, such as the \verb+@rnn+ function in
Listing~\ref{code:rnn_relay}, often certain tensor operators are not
part of the sequential dependency induced by the recursion. For
example, the linear transformation of the input on
line~\ref{line:input_transform} in Listing~\ref{code:rnn_relay} can be
hoisted out of the recursion. Instead of relying on a runtime
scheduling algoritm to identify this as is done in past work,
\Sys~statically discovers such operators that can be hoisted. We
achieve this by relying on a 1-context sensitive taint analysis to
statically compute depths of such operators. We see, in
Listing~\ref{code:rnn_aot}, how the invocation of the kernel
\texttt{bias\_dense} on line~\ref{line:inp_linear_aot} is assigned a
statically computed depth of 0. During runtime, such operators are
thus effectively hoisted out of the recursion. For the RNN example,
this allows us to batch the linear transformations for all input word
embeddings together rather than execute them one at a time.

\subsection{Grain Size Coarsening}\label{sec:ap_coarsen}
\begin{figure}
  \centering \includegraphics[width=0.99\columnwidth]{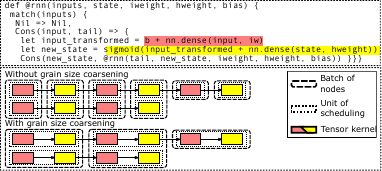}
  \vspace{-2mm}
  \caption{Grain size coarsening for the \texttt{@rnn} function in
    Listing~\ref{code:rnn_relay}.}
  \label{fig:ap_coarsen}
  \vspace{-3mm}
\end{figure}

Generally, scheduling is performed at the granularity of individual
tensor operators i.e. each node in the DFG corresponds to one tensor
kernel call. We saw in \S\ref{sec:survey}, how DL computations
frequently contain larger static sub-graphs embedded in the dynamic
control flow. Therefore, \Sys~performs scheduling at the coarser
granularity of static sub-graphs, thus reducing scheduling
overheads. As these blocks do not contain any control flow, coarsening
the granularity this way does not lead to a loss of exploited
parallelism. This optimization has also been explored in past
work~\cite{jit_batch, cavs, cortex, batchmaker, ebatch} and is
illustrated in Fig.~\ref{fig:ap_coarsen}.

\subsection{Combating Eagerness of Depth Scheduling}
\label{sec:ap_eager}
We saw in \S\ref{sec:inline_depth} how \Sys~relies on ghost operations
and program phases to combating eagerness of depth scheduling. Below,
we provide more detailed explanation of the same.

\noindent\emph{Ghost Operators:} In upper panes of
Fig.~\ref{fig:ghost_ops}, we see that eager batching leads to a
sub-optimal batching schedule in the presence of a conditional
statement. Specifically, the instances of operator B for inputs Inp1
and Inp2 are batched eagerly and, more importantly, separately from
the instances of operator B for inputs Inp3 and Inp4. In the lower
panes, we insert a call to a ghost operator leading to an optimal
schedule. \Sys~statically identifies such cases and insert ghost
operators as needed. Note that ghost operators merely affect
scheduling and are ignored during kernel execution.

\noindent\emph{Program Phases:} For our RNN example in
Listing~\ref{code:rnn_relay}, in order to exploit the most parallelism
for the output operator on line~\ref{line:output_op}, one should wait
until all the operators invoked in the \verb+@rnn+ functionhave been
executed for all the input instances. This way, all output operators
corresponding to all words in all input instances can be executed as
one batched kernel invocation. This would require that all these
output operators be assigned the same depth. However, this may not be
the case as the length of each input sentence may vary. Semantically,
we can divide the RNN computation into two semantic stages---the
initial recursive computations, and the following output
transformations. Given such program phases, \Sys~schedules and
executes operators in one phase before moving on to the next. This
way, \Sys~ensures that all the RNN functions are executed for all
input instances before moving on to the output operators.

\section{More Details on \Sys's Tensor Kernel Generation}\label{sec:ap_kernels}
\subsection{Exploiting Data Reuse}\label{sec:ap_reuse}
\noindent\textbf{Code Duplication for Better Data Reuse:} Code reuse
in the input program can often prohibit the parameter reuse mentioned
above. Consider the following code listing, where, similar to the RNN
model implemented in Listing~\ref{code:rnn_relay}, we implement a
bidirectional RNN (BiRNN)~\cite{birnn} computation. Here, we invoke
the same \verb+@rnn+ function with different model parameters to
implement the forward and backward RNNs. In this case, the tensor
operators invoked by the \verb+@rnn+ function will not be statically
determined to have any arguments constant across multiple calls,
thereby precluding data reuse for the model parameters. In order to
remedy this, before generating the batched kernels, \Sys~recognizes
such cases of data reuse (again using a context-sensitive taint
analysis) and transitively duplicates the necessary functions to
enable data reuse later when generating the batched
kernels\footnote{Simply inlining the \texttt{@rnn} function will not
work here as it is a recursive function.}. In the case of the BiRNN
example, for instance, \Sys~will transitively duplicate the
\verb+@rnn+ function (including the tensor operators it invokes) and
use a different copy of the \verb+@rnn+ function for each of the two
forward and backward calls in the listing below.

\begin{minted}[escapeinside=||,linenos,numbersep=4pt,frame=lines,fontsize=\tiny]{ocaml}
(* Type annotations are omitted in the listing for simplicity. *)
def @main(f_rnn_bias, f_rnn_i_wt, f_rnn_h_wt, f_rnn_init,
          b_rnn_bias, b_rnn_i_wt, b_rnn_h_wt, b_rnn_init,
          inps_list) {
  let rinps_list = @reverse_list(inps_list);
  let forward_res = @rnn(inps_list, f_rnn_init,
                         f_rnn_bias, f_rnn_i_wt, f_rnn_h_wt);
  let backward_res = @rnn(rinps_list, b_rnn_init,
                          b_rnn_bias, b_rnn_i_wt, b_rnn_h_wt);
}
\end{minted}

\begin{figure}
  \centering
  \includegraphics[width=0.99\columnwidth]{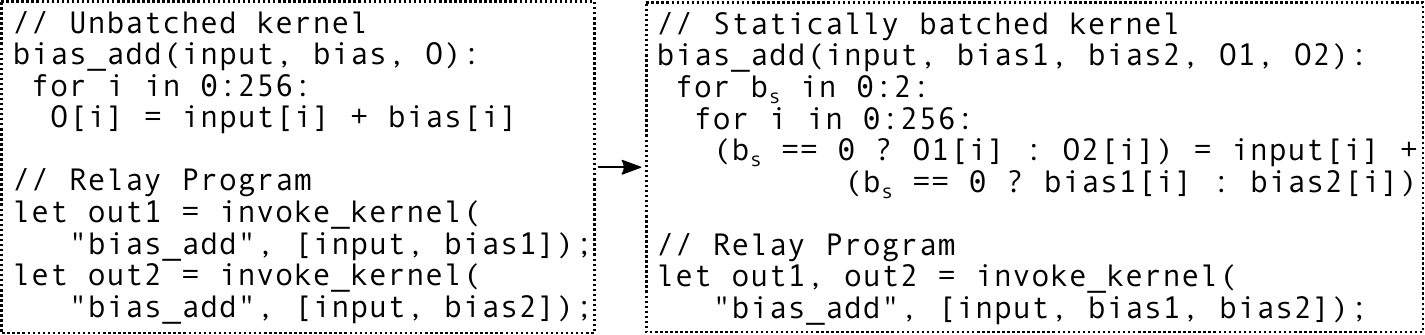}
  \caption{Horizontal fusion promotes parameter reuse.}
  \label{fig:ap_hor_fusion}
\end{figure}

\noindent\textbf{Reuse Within Static Blocks:} Given a tensor operator,
the analysis discussed above takes into account parameters shared
across calls made by different input instances in the mini-batch. This
usually applies to model parameters as they are shared across multiple
input instances. It is often the case, however, that multiple calls to
the same tensor operator within the same static block share a
parameter. For example, this is the case in the commonly used LSTM
cell, where the computation of the four gates all involve concurrent
linear transformations of the same input vector. In such cases,
\Sys~horizontally fuses such calls in order to exploit the parameter
data reuse. This is illustrated in Fig.~\ref{fig:ap_hor_fusion}.

\section{More Implementation Details}
\label{sec:ap_impl}

\subsection{Tensor Kernel Optimization}\label{sec:kernel_opt}
Below, we discuss how \Sys~relies on TVM's auto-scheduler~\cite{ansor}
to automatically generate optimized implementations of batched
versions of (potentially fused) tensor operators used in the input
program.

\noindent\textbf{Auto-scheduler Operator Priorities:} Given a DL
computation consisting of a number of tensor operators, the
auto-scheduler prioritizes the optimization of tensor operators based
on their relative estimated execution cost. Among other factors, this
estimated cost is proportional to the number of times the operator is
invoked during the execution of the input program. In order to
accurately estimate this execution frequency for a given operator in
the presence of control flow (such as repetitive or conditional
control flow), \Sys~relies on profile-guided optimization (PGO). When
PGO is not possible, \Sys~also provides a simple static analysis to
heuristically perform this estimation based on how deeply nested an
operator call is in the recursion.

\noindent\textbf{Handling Variable Loop Extents:} Due to the dynamic
nature of \Sys's scheduling, the loop corresponding to the batch
dimension in the generated unoptimized batched kernels has a variable
extent (kernel \circled{3} in Fig.~\ref{fig:overview}, for
example). In order to optimize these kernels, \Sys~auto-schedules a
corresponding kernel with a static loop extent for the batch dimension
and automatically applies the generated schedule to the original
kernel with the variable extent. Further, when generating code for
loops with variable extents, we often have to insert conditional
checks in order to avoid out of bounds accesses. We rely on the local
padding and local partitioning techniques proposed in
DietCode~\cite{dietcode} to eliminate these conditional checks when
appropriate as they can be severely detrimental to performance

\subsection{Ahead-of-time Compilation}
We saw in \S\ref{sec:impl} that \Sys~compiles the input Relay
computation to C++ in an ahead-of-time fashion. As part of this
compilation, \Sys~lowers all dynamic control flow as well as irregular
data structures to native C++ control flow and classes. Relay handles
scalars by modeling them as zero dimensional tensors. \Sys's AOT
compiler lowers such zero-dimensional tensors and common arithmetic
operators on them to native C++ scalars as well. We see, in
\S\ref{sec:ap_aot_eval}, that this AOT compilation significantly
reduces the execution overheads of dynamic control flow.

\subsection{Other Details}
As discussed in \S\ref{sec:impl} of the main text, we prototype
\Sys~by extending TVM. We find that TVM's operator fusion pass is
limited and is often unable to fuse memory copy operators such as
tensor reshape, concatenation and transpositions. Therefore, in our
implementations of the DL computations, we manually provide fusion
hints to the compiler to force the fusion of such operators with their
consumers. Further, our current prototype only supports the functional
subset of Relay. Specifically, side-effects via mutable references are
currently not supported. \Sys's runtime system has been heavily
optimized to reduce runtime overheads. We use arena allocation (both
on the CPU as well as on the GPU) and asynchronous execution on the
GPU. We also batch memory transfer operations between the CPU and GPU
when possible to reduce the CUDA API overheads.

\section{Supplementary Evaluation and Additional Details}
\begin{table}[t]
  \centering
  \scriptsize
  \caption{NestedRNN (small, batch size 8) execution times
    (without/with PGO), illustrating the benefits of using PGO
    invocation frequencies during auto-scheduling.}
  \addtolength{\tabcolsep}{-3pt}
\resizebox{0.99\columnwidth}{!}{%
  \begin{tabular}{l | ccccc}
    \toprule
    Auto-scheduler iters. & 100 & 250 & 500 & 750 & 1000 \\
    \midrule
Execution times (ms) & 41.08/42.49 & 34.58/30.88 & 31.61/24.4 & 27.33/23.72 & 25.63/24.34 \\
    \bottomrule
  \end{tabular}%
}
  \addtolength{\tabcolsep}{3pt}
  \label{table:pgo_eval}
\end{table}%

\subsection{Benefit of PGO in Tensor Kernel Auto-Scheduling}
\label{sec:ap_pgo_eval}
We mentioned in \S\ref{sec:kernel_opt} that \Sys~uses invocation
frequencies (obtained via PGO) to prioritize tensor operator
optimization during auto-scheduling. In order to evaluate the benefit
of this optimization, we look at the performance of NestedRNN with and
without the optimization. This benchmark computation executes 30
iterations of the inner RNN loop per iteration of the outer GRU loop
on an average. Therefore, the operators invoked in the RNN loop affect
the performance of the benchmark much more than those invoked in the
GRU loop. Table~\ref{table:pgo_eval} shows the execution times of the
benchmark with and without PGO for different iterations of the
auto-scheduler\footnote{Due to the inherent randomness in the
auto-scheduling process, the given execution times are averaged over
10 runs of the auto-scheduler each.} which shows how \Sys~can better
prioritize auto-scheduling for the RNN operators with PGO turned on.